\UseRawInputEncoding
\documentclass{article}
\usepackage[utf8]{inputenc}
\PassOptionsToPackage{numbers,sort&compress}{natbib}
\usepackage[final,main]{neurips_2026}
\usepackage{graphicx}
\usepackage{subcaption}
\usepackage{wrapfig}
\usepackage{multirow}
\usepackage{fvextra}

\usepackage[most]{tcolorbox}
\usepackage[T1]{fontenc}
\usepackage{array}
\newcolumntype{C}[1]{>{\centering\arraybackslash}p{#1}}
\usepackage{svg}
\usepackage{enumitem}
\usepackage[utf8]{inputenc} 
\usepackage{hyperref}       
\usepackage{url}            
\usepackage{makecell}      
\usepackage{booktabs}       
\usepackage{amsfonts}       
\usepackage{nicefrac}       
\usepackage{microtype}      
\usepackage{xcolor}         
\usepackage[capitalize]{cleveref}
\usepackage{amsmath}
\hypersetup{
    colorlinks,
    linkcolor={red!50!black},
    citecolor={blue!50!black},
    urlcolor={blue!80!black}
}

\title{Learning Sampling Parameters for Diffusion Models}

\author{%
  Arisrei Lim \\
  Reve
  \And
  Yossi Gandelsman \\
  Reve
}

\begin{document}

\maketitle

\newcommand{\todo}[1]{\textcolor{red}{#1}}
\newcommand{\whoops}{Whoops-Hard $\cup$ ContraBench}
\newcommand{\method}[0]{LeSAMP}
\newcommand{\winrate}[2]{%
  \ensuremath{{#1}\,{\scriptstyle \pm\,#2}}%
}

\begin{abstract}

Text-to-image diffusion models expose many inference-time sampling parameters, including prompts, negative prompts, classifier-free guidance scales, and noise schedules. These parameters are typically manually chosen once and then held fixed across prompts and denoising timesteps, even though different prompts and stages of generation can benefit from different parameter values. We introduce \method{}, a framework for learning prompt-conditioned, timestep-varying sampling parameters. We formulate parameter selection as a reinforcement learning problem: Given a user prompt, a large language model is trained to emit schedules for the chosen sampling parameters. We optimize our model using rewards from human preference models and VLM-as-a-judge. We evaluate our model on \textsc{Flux.1 [dev]} and Stable Diffusion 3.5, and find that compared to baselines, \method{} has a win rate of up to 68.12\% using human preference scores and 73.37\% using VLM-as-a-judge. These gains are validated in a user study where we achieve win rates of up to 59.46\% over previous baselines. Our results suggest that learned sampling-parameter policies provide a complementary approach to existing post-training methods for improving diffusion model outputs.

\end{abstract}

\section{Introduction}
Diffusion and flow-matching models~\citep{ho2020denoisingdiffusionprobabilisticmodels, song2021scorebasedgenerativemodelingstochastic, lipman2023flowmatchinggenerativemodeling,rombach2022highresolutionimagesynthesislatent} have recently become the dominant paradigm for high-fidelity image generation. Progress has come from novel scalable denoising architectures~\citep{peebles2023scalablediffusionmodelstransformers}, large-scale training on paired text-to-image data~\citep{schuhmann2022laion5bopenlargescaledataset}, and advances in inference techniques and samplers~\citep{song2021scorebasedgenerativemodelingstochastic, lu2022dpmsolverfastodesolver,ho2022classifierfreediffusionguidance}. All these advancements, particularly those at the inference stage, require careful tuning of the hyperparameters (e.g., the sampling parameters of the diffusion process).

While much of the progress in image generation has focused on the training stage and the sampler itself, less attention has been paid to the \textit{choice of sampling parameters}. Modern diffusion models expose a growing hyperparameter space, including classifier-free guidance scale, the noise schedule and step count, and negative prompts. These parameters can interact with text prompts in varying ways. Specifically, different text prompts can benefit from different parameters, and the same prompt can benefit from varying parameters throughout the denoising trajectory \citep{kynkaanniemi2024applying, papalampidi2025dynamicclassifierfreediffusionguidance, huberman2026imagegenerationcontextuallycontradictoryprompts}. In practice, however, sampling parameters are typically hand-picked once per model and held fixed across all prompts, leaving a substantial gap between default and prompt-optimal sampling parameters.

In this paper, we introduce \method{} (\textbf{Le}arning \textbf{Sam}pling \textbf{P}arameters for Diffusion Models), a method that aims to automatically predict prompt-conditioned sampling parameters for text-to-image diffusion models. Concretely, we learn policies that predict four schedulable parameters across denoising timesteps: positive prompt, negative prompt, classifier-free guidance scale, and noise schedule. Rather than using a single fixed value or schedule, we learn to assign different parameter values to different timesteps and prompts.

We formulate each sampling parameter prediction task as a reinforcement learning (RL) problem and fine-tune a large language model (LLM) with GRPO~\citep{shao2024deepseekmathpushinglimitsmathematical} to emit parameter schedules conditioned on the prompt. We investigate two reward signals: a learned human preference model~\citep{ma2025hpsv3widespectrumhumanpreference} and a VLM-based critic that performs pairwise comparison over generated images. Compared to RL post-training of diffusion weights, this design has two attractive properties. First, the diffusion model remains frozen, so its base distribution is preserved and the policy is structurally limited in how far it can drift from the model's natural output, mitigating the mode-collapse and reward-hacking failures common to RL. Second, each policy update only requires running the diffusion sampler in inference mode, removing the need for backpropagation through the denoising chain.

We evaluate \method{} on \textsc{Flux.1~[dev]} (Flux)~\citep{flux2024} and Stable Diffusion 3.5 (SD 3.5)~\citep{esser2024scalingrectifiedflowtransformers}. Against each model's default sampling parameters, SAP~\citep{huberman2026imagegenerationcontextuallycontradictoryprompts}, and R2F~\citep{park2025raretofrequentunlockingcompositionalgeneration}, \method{} trained with VLM-as-a-judge leads to a win rate of up to 68.12\% on human preference scores and 73.37\% with VLM-as-a-judge. These gains are validated in a user study where we achieve win rates up to 59.46\% over SAP and baseline SD 3.5.

Our results suggest that predicting per-step sampling parameters is a simple, model-agnostic, and complementary axis to existing post-training methods for improving diffusion model outputs.

To summarize, our contributions are:
\begin{enumerate}[leftmargin=*]
    \item We introduce \method{}, a method for automatically predicting prompt-conditioned, timestep-varying sampling parameters for text-to-image diffusion models by fine-tuning an LLM policy with GRPO.
    
    \item We define a unified action space over the positive prompt, negative prompt, CFG scale, and noise schedule, with policy-determined span granularity.
    
    \item We show empirically that \method{} improves both human and VLM preference win rates over existing baselines on two diffusion backbones and two reward signals.
\end{enumerate}

\begin{figure}
    \centering
    \includegraphics[width=1\linewidth]{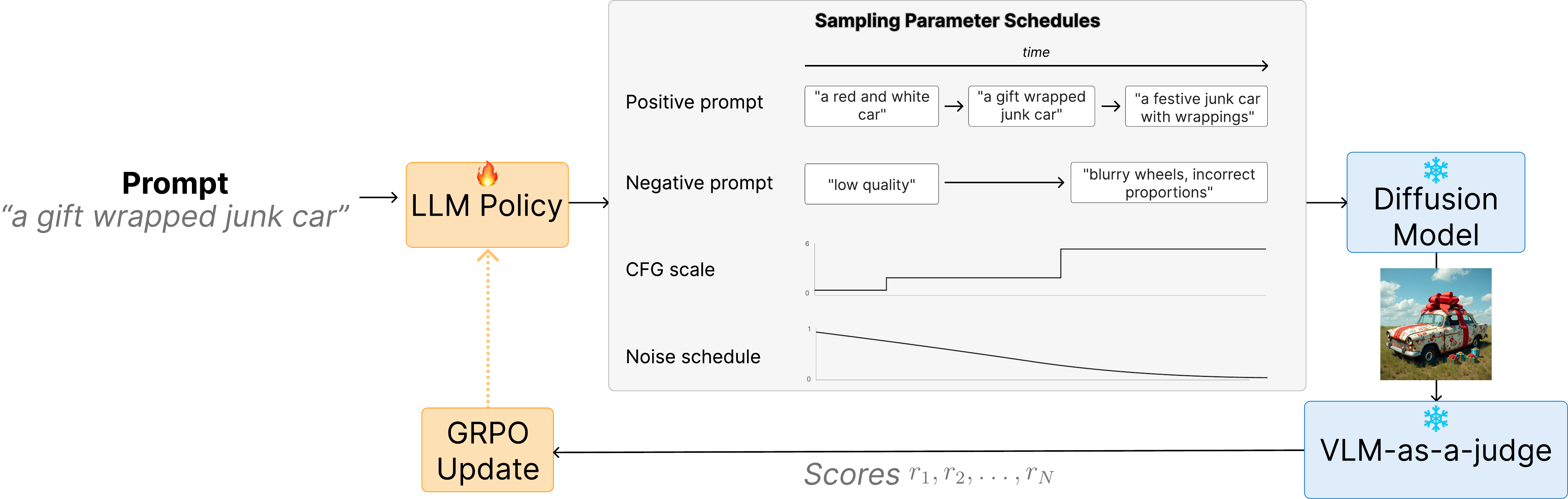}
    \caption{\textbf{\method{} optimization overview.} Using GRPO and VLM-as-a-judge preference reward, we train an LLM policy to predict timestep-varying positive prompt, negative prompt, CFG scale, or noise schedule given a user prompt.}
    \label{fig:method}
\end{figure}

\section{Related Work}

\paragraph{Diffusion models.}
Diffusion models define a generative process by reversing a gradual noising process. Given clean data $x_0 \sim p_{\mathrm{data}}$, the forward process constructs noisy latents $q(x_t \mid x_0)$ by adding Gaussian noise according to a prescribed noise schedule, so that large $t$ corresponds to nearly Gaussian noise. A denoising network is then trained to predict either the added noise, the clean sample, or an equivalent velocity/score parameterization. In the common noise-prediction formulation, the training objective is
\begin{equation}
\mathcal{L}_{\mathrm{DM}}(\theta)
=
\mathbb{E}_{x_0,\,t,\,\epsilon}
\left[
\left\|
\epsilon -
\epsilon_\theta(x_t, t, c)
\right\|_2^2
\right],
\qquad
x_t = \alpha_t x_0 + \sigma_t \epsilon,\quad
\epsilon \sim \mathcal{N}(0,I),
\end{equation}
where $c$ denotes optional conditioning information such as a text prompt, and $(\alpha_t,\sigma_t)$ are determined by the noise schedule~\citep{ho2020denoisingdiffusionprobabilisticmodels,song2021scorebasedgenerativemodelingstochastic}. At inference time, sampling starts from Gaussian noise and repeatedly applies the learned denoiser to produce progressively cleaner samples, using a chosen numerical sampler such as DDIM, DPM-Solver, or an Euler solver for flow-based models~\citep{song2022denoisingdiffusionimplicitmodels,lu2022dpmsolverfastodesolver,lipman2023flowmatchinggenerativemodeling}.

\paragraph{Tuning diffusion sampling parameters.}
The trained denoiser does not by itself fully specify the generation procedure. The quality and behavior of samples also depend on inference-time choices such as the sampler, noise schedule, step count, guidance scale, positive prompt, and negative prompt. Prior work has also found that different denoising timesteps benefit from different sampling behavior. \citet{kynkaanniemi2024applying} show that classifier-free guidance is harmful at high noise levels and unnecessary at low ones, and propose a fixed middle guidance interval. However, these schedules are static across prompts. 

 Papalampidi et al.~\citep{papalampidi2025dynamicclassifierfreediffusionguidance} perform a greedy \textit{per-prompt} search over CFG scales at each timestep using small latent-space evaluators. Unlike our policy, their method does an online search per sample, is limited to CFG, and requires an inference-time evaluator that scores partial latents. Another work, TPDM \citep{ye2025scheduleflydiffusiontime}, optimizes the timestep schedule from latent trajectory features.

\paragraph{RL post-training of diffusion models.} A large body of work fine-tunes diffusion model weights with RL to optimize non-differentiable rewards. DDPO \citep{black2024trainingdiffusionmodelsreinforcement} reframes denoising as a multi-step MDP and applies PPO-style policy gradients, while DPOK \citep{fan2023dpokreinforcementlearningfinetuning} adds KL regularization to a pretrained reference. ReFL \citep{xu2023imagerewardlearningevaluatinghuman} and DRaFT \citep{clark2024directlyfinetuningdiffusionmodels} instead backpropagate differentiable rewards through a truncated portion of the sampling chain. Preference-based variants such as Diffusion-DPO \citep{wallace2023diffusionmodelalignmentusing} extend DPO \citep{rafailov2024directpreferenceoptimizationlanguage} to image generation, and Flow-GRPO \citep{liu2025flowgrpotrainingflowmatching} adapts GRPO \citep{shao2024deepseekmathpushinglimitsmathematical} to flow-matching models by converting the deterministic ODE into an SDE for exploration. All of these methods modify diffusion weights, and some also require backpropagation through the sampling trajectory. Our method is complementary. We freeze the diffusion model and train an external LLM policy over its sampling parameters, preserving the base diffusion model's distribution and making each policy update much cheaper.

\paragraph{LLM-guided prompt and sampling control for text-to-image generation.}
A second line of work uses an LLM as an interface for controlling diffusion outputs. Promptist~\citep{hao2023optimizingpromptstexttoimagegeneration} fine-tunes a language model with PPO to rewrite the user prompt into a single, model-preferred form, optimizing a combined aesthetic and CLIP reward. Other training-free methods prompt an LLM to plan time-varying conditioning -- R2F~\citep{park2025raretofrequentunlockingcompositionalgeneration} uses an LLM to alternate between rare concepts in the prompt and surrogate concepts with similar visual features at each denoising step. SAP~\citep{huberman2026imagegenerationcontextuallycontradictoryprompts} targets prompts containing contextual contradictions. It queries an LLM to decompose the target prompt into a sequence of proxy prompts applied at different timesteps. 

\method{} combines complementary aspects of these methods. Like Promptist, we train an LLM with RL to output sampling parameters that optimize an image-level reward, and like R2F and SAP, our policy emits timestep-varying sampling parameters rather than a single static value.  Recent work has also explored automating negative prompts. NPC \citep{park2025guidinggenerateautomatednegative} uses an LLM to propose candidate negative prompts and searches for the best one during inference time. In contrast, our approach uses an LLM to emit conditioning parameters in a single generation, avoiding test-time search. Furthermore, while NPC focuses on static negative prompts, we predict timestep-varying parameter schedules.

\section{Learning to Predict Diffusion Sampling Parameters via RL}
\label{sec:method}
In this section, we present our method, \method, shown in Figure~\ref{fig:method}. We formulate the task of choosing diffusion sampling parameters as a Markov decision process. Then, we utilize reinforcement learning to optimize an LLM to produce these parameters, given a reward function and a prompt. Finally, we discuss different reward functions and the types of images they encourage.

\subsection{Preliminaries}
\label{sec:preliminaries}

\textbf{Classifier-free guidance.} For text-to-image models, sampling is generally conditioned on a prompt. \emph{Classifier-free guidance} (CFG)~\citep{ho2022classifierfreediffusionguidance} sharpens conditioning by extrapolating between a conditional and an unconditional noise prediction:
\begin{equation}
\tilde{\epsilon}_\theta(x_t, c_+, c_-, w)
= \epsilon_\theta(x_t, c_-) + w\bigl(\epsilon_\theta(x_t, c_+) - \epsilon_\theta(x_t, c_-)\bigr),
\label{eq:cfg}
\end{equation}
where $c_+$ is the conditioning (positive) prompt, $c_-$ is an unconditional (negative) prompt, and $w \geq 1$ is the guidance scale. Larger $w$ strengthens prompt conditioning and can improve sample fidelity. However, it also reduces diversity and can introduce artifacts such as oversaturation~\citep{ho2022classifierfreediffusionguidance,lin2024commondiffusionnoiseschedules,sadat2025eliminatingoversaturationartifactshigh}.

\textbf{Noise schedule.} The noise schedule specifies the sequence of noise levels $\sigma_0 > \sigma_1 > \cdots > \sigma_T$ followed by the sampler during denoising. Larger values of $\sigma_t$ correspond to earlier, more corrupted latents, while smaller values correspond to later, cleaner latents. At inference time, the sampler starts from an initial high-noise latent and iteratively moves through this sequence toward $\sigma_T \approx 0$, applying the denoising model
at each step. The choice of schedule determines how the finite sampling budget is allocated across high to low noise regimes, and can affect global structure, fine details, prompt adherence, and overall image quality. Standard text-to-image pipelines fix this sequence in advance, often using a hand-designed rule such as a linear, cosine, Karras, or model-specific default schedule.

\textbf{Policy controls.} We train policies to control each of these diffusion sampling parameters: positive prompt $c_+$, negative prompt $c_-$, CFG scale $w \in [w_{\min}, w_{\max}]$, and noise schedule $\sigma=[\sigma_0, \sigma_1, \dots, \sigma_T]$. While standard practice fixes $c_+$, $c_-$, $w$, and the noise schedule $\{\sigma_t\}$ for the full trajectory, we treat each of these as a \emph{schedulable} parameter that an LLM policy can vary across the denoising trajectory (Section~\ref{sec:param-space}).

\textbf{Group Relative Policy Optimization (GRPO). }
GRPO~\citep{shao2024deepseekmathpushinglimitsmathematical} is a policy gradient method that replaces the learned value function of PPO~\citep{schulman2017proximalpolicyoptimizationalgorithms} with a group-relative baseline. For each prompt $x$ in a training batch, GRPO samples a group of $G$ rollouts $\{o_1, \dots, o_G\}$ from the current policy $\pi_{\theta_{\text{old}}}$ and assigns each a reward $r_i = R(o_i)$. The advantage for rollout $i$ is computed by normalizing within the group:
\begin{equation}
\hat{A}_i = \frac{r_i - \mathrm{mean}(\{r_j\}_{j=1}^G)}{\mathrm{std}(\{r_j\}_{j=1}^G)}.
\end{equation}
The policy is updated by maximizing a PPO-style clipped surrogate objective with a KL penalty to a reference policy $\pi_{\text{ref}}$:
\begin{equation}
\mathcal{J}(\theta) =
\mathbb{E}\!\left[ \frac{1}{G}\sum_{i=1}^{G} \frac{1}{|o_i|}
\sum_{t=1}^{|o_i|} \min\!\bigl( \rho_{i,t}\hat{A}_i,\; \mathrm{clip}(\rho_{i,t}, 1-\epsilon, 1+\epsilon)\hat{A}_i \bigr)
- \beta\, D_{\mathrm{KL}}\!\left[\pi_\theta \,\|\, \pi_{\text{ref}}\right] \right],
\end{equation}
where $\rho_{i,t} = \pi_\theta(o_{i,t} \mid x, o_{i,<t}) / \pi_{\theta_{\text{old}}}(o_{i,t} \mid x, o_{i,<t})$ is the per-token importance ratio, and $\epsilon, \beta$ are hyperparameters controlling clipping and KL regularization. Because our reward is defined only at the end of the episode (after parsing the rollout, running the diffusion model, and scoring the image), every token in a rollout shares the same group-normalized advantage $\hat{A}_i$.

\subsection{Problem Setting}
We formalize our setting as a finite-horizon MDP $(S, A, P, R, \gamma)$. We treat the diffusion model as part of the environment. The LLM policy generates tokens that are parsed into sampling parameters, and the diffusion model is called at the episode end to produce a sample that is then given a reward.

The state space $S$ consists of all possible token sequences in the LLM's context window, so a state $s_t \in S$ corresponds to the original prompt concatenated with all tokens generated up to step $t$. The action space $A$ is the LLM's vocabulary $\mathcal{V}$, and an action $a_t \in A$ is the next token sampled from the policy. The transition kernel $P$ is deterministic. Given $(s_t, a_t)$, $s_{t+1}$ is obtained by appending $a_t$ to $s_t$. The reward $R$ is given by parsing the produced tokens into sampling parameters, calling a diffusion model with these parameters, and scoring the sample with a chosen reward model. Finally, we set $\gamma = 1$, since each episode terminates at the maximum token budget or earlier.

\subsection{Generating Sampling Parameters}
\label{sec:param-space}
\textbf{LLM Inputs. }The LLM is prompted with a description of each parameter and its allowed value range, and responds with a JSON object containing a subset of the four keys. Any parameter not emitted falls back to the diffusion model's default sampling parameters. This design lets us study each parameter in isolation by restricting the prompt to a single key, and also allows the policy to combine them. See Appendix~\ref{sec:policy_prompts} for prompts used.

\textbf{LLM Outputs. }The policy outputs a JSON string that is parsed into a schedule over the denoising trajectory. Let the diffusion model run for $T$ denoising steps. For a sampling parameter $p$, a schedule is a partition of $[0, T]$ into contiguous spans $\{[\tau_0^p, \tau_1^p), [\tau_1^p, \tau_2^p), \dots, [\tau_{K-1}^p, \tau_K^p]\}$ with $\tau_0^p = 0$ and $\tau_K^p = T$, together with a value $v_k^p$ for each span $k$. This allows the policy to use coarse or fine schedules depending on the prompt and token budget. See Appendix~\ref{sec:policy-outputs} for example outputs.

Outputs that fail to parse or contain invalid values (e.g., spans that do not cover $[0, T]$) are assigned a penalty reward relative to the GRPO group's valid rewards. Concretely, for a prompt $x$ with group $\mathcal{G}(x) = \{o_1, \dots, o_G\}$ of $G$ rollouts, let $\mathcal{G}_{\mathrm{valid}}(x) \subseteq \mathcal{G}(x)$ denote the rollouts that parse and produce an image. Let $\sigma_{\mathrm{valid}}$ denote the standard deviation of the rewards assigned to valid rollouts in the group. The reward is
\begin{equation}
R(o_i) =
\begin{cases}
  R_{\mathrm{model}}(\mathrm{image}(o_i)) 
  & o_i \in \mathcal{G}_{\mathrm{valid}}(x), \\[4pt]
  \displaystyle
  \min_{o_j \in \mathcal{G}_{\mathrm{valid}}(x)}
  R_{\mathrm{model}}(\mathrm{image}(o_j))
  - \mathrm{max(}\lambda_{\mathrm{fail}} \sigma_{\mathrm{valid}}, 1\mathrm{)}
  & o_i \notin \mathcal{G}_{\mathrm{valid}}(x).
\end{cases}
\end{equation}
Here $\lambda_{\mathrm{fail}}$ is the failure penalty coefficient, which we set to 2. This ensures that invalid outputs receive a strictly lower reward than any valid output in their group. If no valid completions exist in the group, that prompt is dropped from the training batch.

\subsection{Reward Functions}
\label{method:reward}
In our experiments, we explore a variety of reward functions to guide the LLM to generate sampling parameters that lead to preferred images. These reward functions are used to score each image in a group in our GRPO implementation.

\paragraph{VLM-as-a-judge.}

We use VLM-as-a-judge by prompting a VLM (e.g., Gemini 3 Flash) to compare two images and return the index of the better image. We repeat this for all unique pairs of images in a group, collecting win statistics over each pair. The score of each image is its group-level win rate. More specifically, the VLM is prompted to choose the image that has better adherence to the generation request and better aesthetics. If both images are equally good or equally bad, the model is prompted to return a tie. To prevent bias from image position, we query the VLM twice with both orderings of the images. If any response contains a tie or responses disagree, we consider the final evaluation to be a tie. Unless otherwise stated, the main reward function we use in training is image win rate under Gemini 3 Flash as a critic.

\paragraph{HPSv3.}
HPSv3~\citep{ma2025hpsv3widespectrumhumanpreference}, a human preference model trained on 1.17 million pairwise text-to-image rankings, is another reward function candidate. Given an image and a prompt, it outputs a scalar that aims to represents how well the image aligns with the prompt. Using HPSv3 as a reward signal allows us to verify the extent to which we can learn sampling parameters to generate human-preferred images. We note that HPSv3 is susceptible to reward hacking via its bias toward aesthetic but non-prompt-adherent images. For this reason, we find that HPSv3 is not an optimal reward signal to train on. We show the results of training with HPSv3 and discuss its limitations in Appendix~\ref{sec:hpsv3_experiments}.

\begin{figure}
    \centering
    \includegraphics[width=1\linewidth]{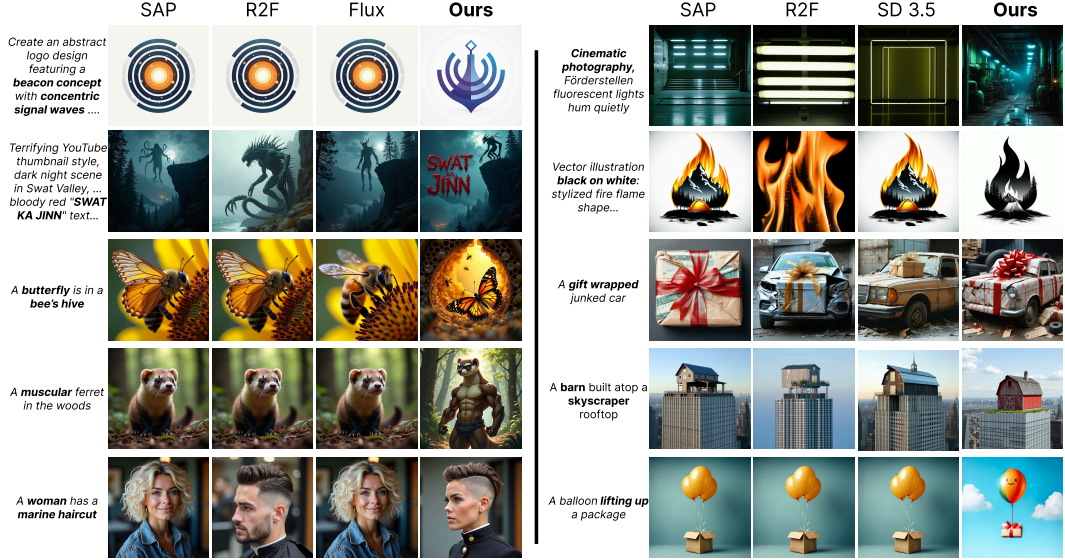}
    \caption{Comparison between \method{}, SAP, and R2F, applied to SD 3.5 and Flux. Each row shows images generated from a prompt shown on the left, using the methods in the columns.}
    \label{fig:baseline_comparison}
\end{figure}

\section{Experiments}
\label{sec:experiments}

We evaluate whether \method{} can improve image aesthetics and prompt adherence over baselines, and discuss the distribution of learned sampling parameters. Our experiments fine-tune Qwen3-32B with GRPO to produce the sampling parameters described in Section~\ref{sec:preliminaries}. We train the policy to maximize VLM-as-a-judge reward, using Gemini 3 Flash. We use Stable Diffusion 3.5~\citep{esser2024scalingrectifiedflowtransformers} and \textsc{Flux.1 [dev]}~\citep{flux2024} as our diffusion backbones. We also show results on Reve 2.0 \citep{reve2026layoutbet} in Appendix~\ref{sec:reve-results}, demonstrating that \method{} generalizes to frontier models. For implementation details, see Appendix~\ref{sec:impl_details}.

We evaluate on the union of Whoops-Hard and ContraBench, which we refer to as \whoops{}. These datasets come from SAP \citep{huberman2026imagegenerationcontextuallycontradictoryprompts} and contain contextually contradictory prompts (e.g., ``A dragon blowing water''). To better capture the distribution of common image generation requests, we also collect and evaluate on 240 user image generation requests from the web, which we refer to as User Requests.

\paragraph{Baselines.}
We compare \method{} against three baselines. First, we use each diffusion model's default sampling parameters, keeping the original prompt fixed. We also compare against SAP~\citep{huberman2026imagegenerationcontextuallycontradictoryprompts}, which uses an off-the-shelf LLM to decompose a contextually contradictory prompt into a sequence of proxy prompts for different denoising stages. Finally, we compare to R2F~\citep{park2025raretofrequentunlockingcompositionalgeneration}, which uses an off-the-shelf LLM to swap rare concepts with visually related frequent concepts, switching back and forth between the original rare prompt and proxy frequent prompts. We replace SAP and R2F's original GPT-4o LLM with GPT-5.4 mini while keeping everything else unchanged. Since GPT-5.4 mini is stronger than off-the-shelf Qwen3-32B, this makes the prompt-scheduling baselines more competitive and ensures our gains come from RL training rather than from using a stronger base LLM.

\begin{table}[t]
\centering
\setlength{\tabcolsep}{8pt}
\begin{tabular}{l c c c c}
\toprule
& \multicolumn{2}{c}{\textbf{\whoops{}}}
& \multicolumn{2}{c}{\textbf{User Requests}} \\
\cmidrule(lr){2-3} \cmidrule(lr){4-5}
\textbf{Comparison}
& Gemini & HPSv3
& Gemini & HPSv3 \\
\midrule

Ours vs SAP$_\text{SD 3.5}$ & \winrate{64.58}{6.23} & \winrate{57.25}{8.25} & \winrate{62.71}{4.93} & \winrate{56.67}{6.27} \\
Ours vs R2F$_\text{SD 3.5}$ & \winrate{67.84}{6.51} & \winrate{68.12}{7.77} & \winrate{68.91}{4.77} & \winrate{61.25}{6.16} \\
Ours vs SD 3.5              & \winrate{66.49}{6.19} & \winrate{60.14}{8.16} & \winrate{60.10}{5.00} & \winrate{52.08}{6.32} \\

\midrule

Ours vs SAP$_\text{Flux}$  & \winrate{67.48}{6.14} & \winrate{51.54}{8.33} & \winrate{64.27}{4.93} & \winrate{64.58}{6.05} \\
Ours vs R2F$_\text{Flux}$  & \winrate{73.37}{5.91} & \winrate{63.04}{8.05} & \winrate{69.27}{4.92} & \winrate{63.75}{6.08} \\
Ours vs Flux               & \winrate{69.38}{5.84} & \winrate{55.80}{8.28} & \winrate{60.31}{4.92} & \winrate{57.50}{6.25} \\

\bottomrule
\end{tabular}
\vspace{0.25em}
\caption{Pairwise win rates. Each cell reports the tie-adjusted win rate $(W + 0.5T) / (W + L + T)$ where $W$, $L$, and $T$ are wins, losses, and ties, respectively. 95\% confidence intervals are reported with $\pm$ margins, using the number of prompts as the effective sample size. Each row is evaluated on the dataset in the top-level column using the reward function in the second-level column.}
\label{tab:baseline-comparisons}
\end{table}

\paragraph{Qualitative Results.}
\label{subsec:qualitative_results}
Figure~\ref{fig:baseline_comparison} presents a qualitative comparison of \method{} to baselines. The first two rows use prompts from our User Requests dataset, and the remaining rows use prompts from \whoops{}. Using the default sampling parameters often creates images that miss important details on User Requests and do not accurately represent the contextually contradictory prompts from \whoops{}. 
SAP and R2F improve some contradictory cases, but because they rely on prompt decomposition or concept substitution, they can degrade ordinary user requests. For example, in the ``Terrifying YouTube thumbnail...'' prompt, both methods introduce the required ``SWAT KA JINN'' text only at later denoising steps, after the image layout has largely formed. As a result, the final image fails to render the text. On contextually contradictory prompts such as ``A gift wrapped junk car,'' SAP and R2F correctly recognize the contradiction, but their prompt-swapping strategies still fail to produce an image that fully captures the intended composition.

Our trained policy takes a different approach. Instead of swapping between common and rare concepts, our policy outputs shorter prompts at the beginning and gradually adds detail in later steps. Importantly, we note that the initial predicted prompt already has a significant amount of detail, and sometimes more detail than in a short original prompt. With these prompt schedules, we observe that our policy leads to images that are both prompt-adherent and aesthetically pleasing.

These findings suggest that \textit{concept-substitution methods are not strictly needed} to achieve good results on contextually contradictory prompts. Rather, gradually adding more detail to prompts throughout the denoising process can satisfy requirements for both contextually contradictory and in-distribution user requests.

\paragraph{Quantitative Results.}

As shown in Table~\ref{tab:baseline-comparisons}, across each dataset and evaluation metric, \method{} has a win rate greater than 50\%, suggesting that \method{} outperforms all baselines. We note that while our policy was trained to maximize Gemini preference win rate, it was never trained to maximize HPSv3, reducing the likelihood that our results stem from unwanted reward hacking. In Table~\ref{tab:sap-scores}, we also replicate the prompt adherence and image quality scoring method used by SAP, but switch to a stronger model, GPT-5.4 mini. Under these metrics, we continue to see that \method{} outperforms or matches prior methods on both alignment and quality.

\paragraph{User Study.}
To validate our previous results, we conduct a user study comparing \method{}, SAP, and default sampling parameters. Users compare two images for each prompt, considering prompt adherence and image aesthetics. Across \whoops{} and User Requests, 13 evaluators provide 511 pairwise preference labels. As shown in Table~\ref{tab:user-study}, we see that images from \method{} are preferred over those from SAP and SD 3.5, supporting our previous findings.



\begin{table}[t]
\centering
\begin{tabular}{llc}
\toprule
\textbf{Dataset} & \textbf{Comparison} & \textbf{Win rate} \\
\midrule
\whoops{} & Ours vs $\text{SAP}_{\text{SD }3.5}$ & 53.68 \\
\whoops{} & Ours vs SD 3.5 & 59.46 \\
User Requests & Ours vs $\text{SAP}_{\text{SD }3.5}$   & 54.29 \\
User Requests & Ours vs SD 3.5 & 55.81 \\
\bottomrule
\end{tabular}
\vspace{0.25em}
\caption{Pairwise preference data collected from a user study reporting tie-adjusted win rates.}
\label{tab:user-study}
\end{table}

\begin{table}[t]
\centering
\begin{tabular}{llccc}
\toprule
Dataset & Metric & Ours & $\text{SAP}_{\text{SD }3.5}$ & $\text{R2F}_{\text{SD }3.5}$ \\
\midrule
\multirow{2}{*}{\whoops{}}
  & SAP Alignment & \textbf{4.07} & 3.76 & 3.76\\
  & SAP Quality   & \textbf{4.68} & \textbf{4.68} & 4.62\\
\midrule
\multirow{2}{*}{User Requests}
  & SAP Alignment & \textbf{4.07} & 3.94 & 3.65\\
  & SAP Quality   & \textbf{4.58} & 4.57 & 4.50\\
\bottomrule
\end{tabular}
\vspace{0.25em}
\caption{SAP Alignment and Quality scores evaluated on \method{}, SAP, and R2F.}
\label{tab:sap-scores}
\end{table}

\begin{table}[t]
\centering
\begin{tabular}{@{}l@{\hspace{1.5em}}l@{\hspace{1.5em}}c@{\hspace{1.5em}}c@{}}
\toprule
\textbf{Diffusion model} & \textbf{Trained parameter}
& \textbf{Gemini}
& \textbf{HPSv3} \\
\midrule

Stable Diffusion 3.5 & Positive prompt & 60.10 & 52.08 \\
Stable Diffusion 3.5 & Negative prompt & 56.46 & 58.75 \\
Stable Diffusion 3.5 & CFG Scale       & 54.79 & 53.75 \\
Stable Diffusion 3.5 & Noise Schedule  & 51.46 & 48.33 \\

\midrule

Flux.1 [dev] & Positive prompt & 60.31 & 57.50 \\
Flux.1 [dev] & Negative prompt & 62.29 & 35.00 \\
Flux.1 [dev] & CFG Scale       & 48.85 & 58.75 \\
Flux.1 [dev] & Noise Schedule  & 51.04 & 51.25 \\

\bottomrule
\end{tabular}
\vspace{0.25em}
\caption{
    Win rates against default sampling parameters when using Gemini-as-a-judge preference or HPSv3 score after training with Gemini preference reward, evaluated on User Requests.
}
\label{tab:param-ablation}
\end{table}

\begin{figure}
    \centering
    \includegraphics[width=1\linewidth]{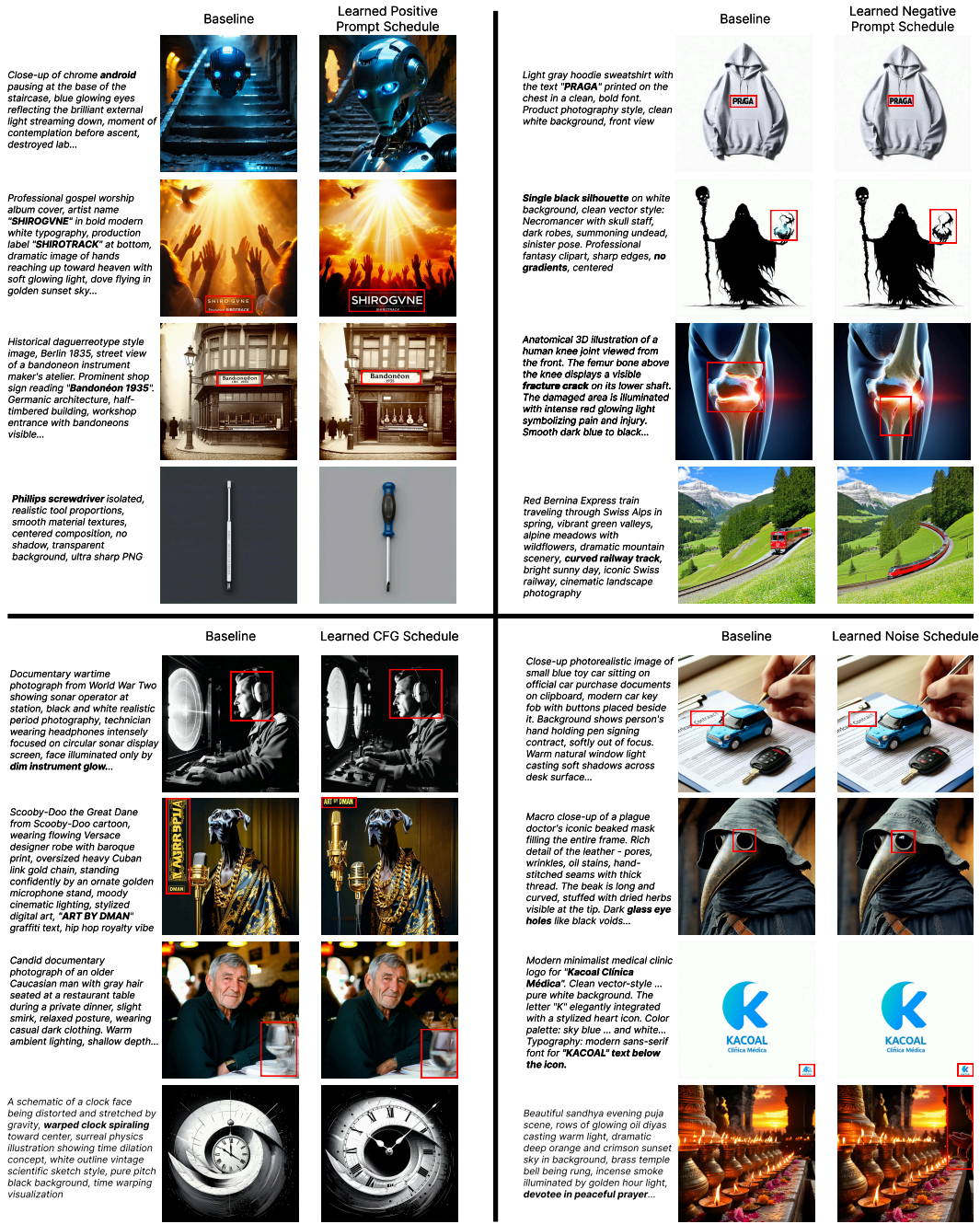}
    \caption{
    Results from four \method{}-trained policies predicting schedules for positive prompt, negative prompt, CFG, and noise on SD 3.5. Areas of interest are boxed in red. Images without boxes indicate that the global context is of interest. Please zoom in for the best details.
    }
    \label{fig:sd35-param-ablation}
\end{figure}

\subsection{Ablations}

\paragraph{Effect of predicting each sampling parameter.}
In Table~\ref{tab:param-ablation} and Figure~\ref{fig:sd35-param-ablation}, we show the results of training our policy to predict each sampling parameter in Section~\ref{sec:preliminaries}, and evaluate using the reward functions in Section~\ref{method:reward} on Flux and SD 3.5. These results indicate that our policy is able to output positive and negative prompts that outperform the baseline across nearly all settings. Detailed examples of outputs are shown in Appendix~\ref{sec:policy-outputs}.

Across both diffusion models, we find that our policy learns to output progressively longer prompts at later denoising steps. More detail about this behavior is discussed in the Qualitative Results. Negative prompts behave differently between Flux and SD 3.5. In SD 3.5, the policy has learned a mostly universal negative prompt schedule, cycling through generic photographic-quality terms (e.g., \emph{blurry, low resolution}) at intermediate and late steps. In Flux, the negative prompt at the first and last denoising steps is mostly fixed, similar to SD 3.5. However, intermediate steps adapt to the prompt content (e.g., \emph{blurry faces} on portrait prompts).

With CFG and noise schedule, we see less consistent improvements. In SD 3.5, we do observe an improvement from predicting CFG scale. Here, our policy has learned to output a monotonically increasing CFG schedule, starting near 4 and ending around 8. However, we do not see the same improvement in Flux. We believe this is because Flux is guidance-distilled, resulting in a less rich and harder-to-predict action space. For noise schedule, we observe that across both models, our policy learns to predict an approximately linearly decreasing schedule, similar to the default schedule. Given that our win rates with noise schedule are $\sim$50\%, this behavior makes sense.

Finally, we observe that positive prompt gives the best combination of gains on the training reward and on a held-out human preference model. For this reason, we choose to predict the positive prompt for the majority of our experiments.

\paragraph{Effect of timestep-varying sampling parameters.} To isolate the effect of predicting time-varying positive prompts as opposed to a single prompt, we train a policy to predict the positive prompt for SD 3.5 without changing it at different timesteps, similarly to Promptist~\citep{hao2023optimizingpromptstexttoimagegeneration}. While we observe a Gemini-judge win rate of 55.62\%, indicating improvement over the default sampling parameters, this win rate is still lower than the 60.10\% win rate we observe when predicting timestep-varying positive prompts in Table~\ref{tab:param-ablation}. This gap isolates the benefit of timestep variation, suggesting that the gains are not solely due to generic prompt optimization, but also to scheduling the conditioning over time. Both policies can rewrite the user prompt, but only \method{} can change the conditioning over the denoising trajectory.

\section{Discussion, Limitations, and Future Work}
\label{sec:discussion}

Our results suggest that learning prompt-conditioned sampling parameters is an effective way to improve text-to-image generation without modifying the underlying diffusion model. Across diffusion backbones, datasets, and evaluation metrics, \method{} improves over default sampling parameters and prior LLM-based prompt-scheduling baselines. 

This indicates that a substantial part of the gap between default and preferred generations can be closed through better inference-time conditioning.

\paragraph{Limitations and future work.}
We find that the choice of reward model is critical. When training directly with HPSv3, we observe severe reward hacking. The generated images are often aesthetically strong in isolation, but no longer satisfy the user prompt. We elaborate on our reward hacking observations in Appendix~\ref{sec:hpsv3_experiments}.

We also find that jointly training over positive prompts, negative prompts, CFG schedules, and noise schedules does not substantially improve over training on positive-prompt-only. The jointly trained policy achieves similar win rates to the positive-prompt-only policy, with a 60.42\% Gemini win rate and a 54.58\% HPSv3 win rate against the default sampler. One possibility is that the larger action space requires more exploration. Another possibility is that the most accessible gains for our models and datasets already come from time-varying positive prompts. We believe that investigation in both areas can reveal more information about how to best set diffusion model sampling parameters.

Finally, policies trained with our method still fail on some difficult prompts. Because our training distribution consists of ordinary user requests on which such failures are rare, one future direction could explore automatically generating large sets of difficult user requests, which could then be used as training data within our framework.

\begin{ack}
We thank Taesung Park for coming up with the initial idea of learning prompt-conditioned sampling parameters.

\end{ack}

\newpage
\bibliography{neurips_2026}
\bibliographystyle{unsrtnat}

\newpage
\appendix

\begin{figure}
    \centering
    \includegraphics[width=1\linewidth]{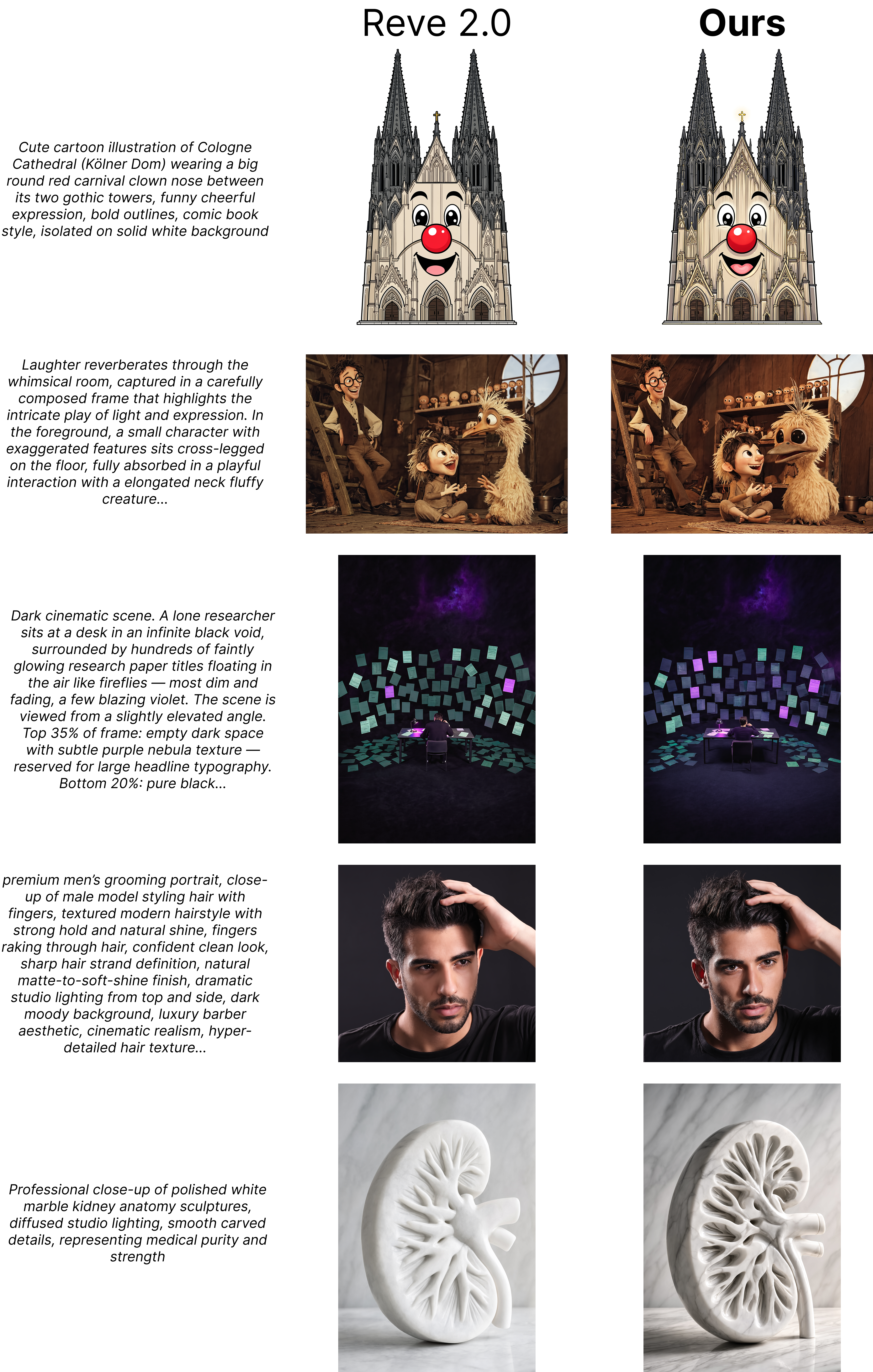}
    \caption{Comparison between Reve 2.0 and \method{} applied to Reve 2.0. Please zoom in for the best details.}
    \label{fig:reve_comparisons}
\end{figure}

\begin{table}[t]
\centering
\begin{tabular}{llc}
\toprule
\textbf{Dataset} & \textbf{Comparison} & \textbf{Win rate} \\
\midrule
\whoops{} & Ours vs Reve 2.0 & 55.00 \\
User Requests & Ours vs Reve 2.0 & 68.00 \\
\bottomrule
\end{tabular}
\vspace{0.25em}
\caption{Pairwise preference data collected from a user study. We show tie-adjusted win rates of \method{} against Reve 2.0, evaluated on User Requests.
}
\label{tab:reve-user-study}
\end{table}

\begin{figure}
    \centering
    \includegraphics[width=0.5\linewidth]{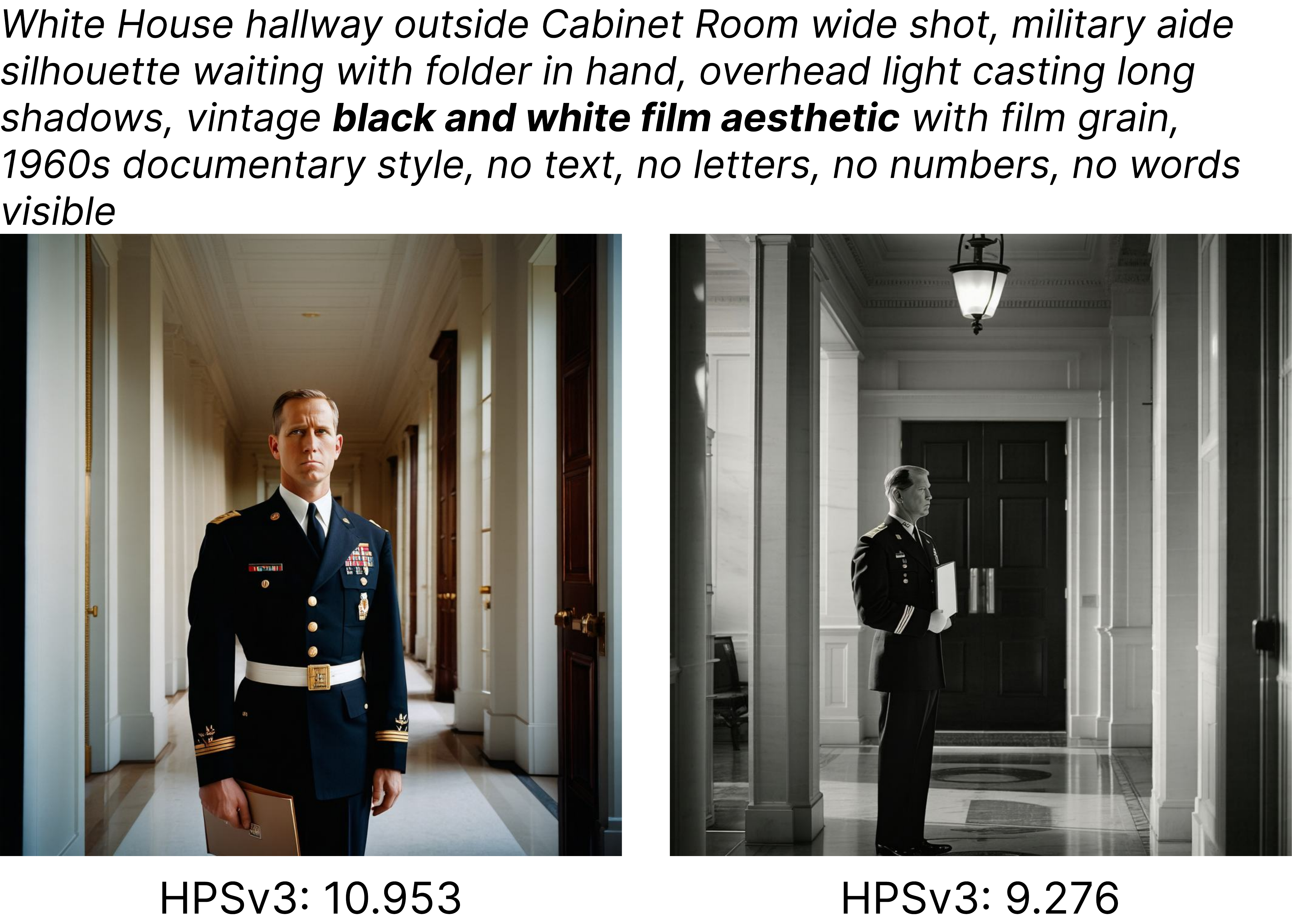}
    \caption{Example of HPSv3's bias toward aesthetic but non-prompt-adherent images. The left image scores higher than the right image, despite obviously not reflecting what the prompt describes.}
    \label{fig:hpsv3-reward-hacking}
\end{figure}

\begin{table}[t]
\centering
\begin{tabular}{@{}l@{\hspace{1.5em}}l@{\hspace{1.5em}}c@{\hspace{1.5em}}c@{}}
\toprule
\textbf{Diffusion model} & \textbf{Trained parameter}
& \textbf{Gemini}
& \textbf{HPSv3} \\
\midrule

Stable Diffusion 3.5 & Positive prompt & 37.67 & 65.42 \\
Stable Diffusion 3.5 & Negative prompt & 57.29 & 73.75 \\
Stable Diffusion 3.5 & CFG Scale       & 42.29 & 63.33 \\
Stable Diffusion 3.5 & Noise Schedule  & 52.71 & 47.92 \\

\midrule

Flux.1 [dev] & Positive prompt & 12.50 & 67.50 \\
Flux.1 [dev] & Negative prompt & 65.79 & 42.50 \\
Flux.1 [dev] & CFG Scale       & 47.40 & 65.00 \\
Flux.1 [dev] & Noise Schedule  & 57.92 & 65.83 \\

\bottomrule
\end{tabular}
\vspace{0.25em}
\caption{
    Win rates against default sampling parameters when using Gemini-as-a-judge preference or HPSv3 score after training with HPSv3 reward.
}
\label{tab:hpsv3-param-ablation}
\end{table}
     
\section{Results on Reve 2.0}
\label{sec:reve-results}
We evaluate \method{} on Reve 2.0~\citep{reve2026layoutbet}, a frontier image generation model whose architecture differs substantially from SD 3.5 and Flux. A natural concern with new approaches is that they mainly compensate for capability gaps in weaker models. However, we observe that \method{} still improves upon baseline Reve 2.0, even though Reve 2.0 is a much stronger model than SD 3.5 and Flux. These findings are validated by a user study, where we ask 10 people to each annotate 50 pairs of Reve 2.0 vs \method{}. In this study, we do not give the option for a tie, so annotators must always pick win or lose. To reduce the effect of low-confidence labels, we treat any pair that has between 4 and 6 wins for either model as a tie. We show tie-adjusted win rates from this study in Table~\ref{tab:reve-user-study}, showing that \method{} is consistently preferred over the baseline in terms of user preference. We also show qualitative results in Figure~\ref{fig:reve_comparisons}.

\section{HPSv3 Experiments}
\label{sec:hpsv3_experiments}
Here, we discuss the performance of policies after training with HPSv3 reward. We also discuss the reward hacking and biases that occur when training and evaluating with HPSv3.

\subsection{HPSv3 Training Results}
In Table~\ref{tab:hpsv3-param-ablation}, we show the results of training each available sampling parameter on SD 3.5 and Flux. Across most sampling parameters we are able to increase HPSv3 win rate at the expense of Gemini win rate. This indicates that the two rewards are not strongly correlated. We discuss a possible reason why in Appendix~\ref{subsec:hpsv3-reward-hacking}.

\subsection{HPSv3 Reward Hacking}
\label{subsec:hpsv3-reward-hacking}
In Figure~\ref{fig:hpsv3-reward-hacking}, we can see an example of HPSv3's bias toward aesthetic but non-prompt-adherent images. The prompt requests a black and white image. The image on the left is in color, and the image on the right is in black and white. However, HPSv3 still gives the non-prompt-adherent left image a higher score than the prompt-adherent right image. When training with HPSv3, this behavior transfers to our policy. It learns to ignore small but important details in the prompt and instead prioritizes absolute aesthetics over prompt adherence. This is especially apparent when training to predict a positive prompt schedule. We hypothesize that the policy's learned behavior stems from bias in HPSv3.

However, if we assume images are prompt adherent, HPSv3's bias can be neutralized. Then, we can use HPSv3 purely for evaluating image aesthetics. One possible direction to explore involves a ``VLM-gated'' HPSv3, where a VLM is used to first decide prompt adherence and HPSv3 is used to score aesthetics.

\section{User Study}
\label{sec:user_study_details}
Users were compensated for this study as part of their regular job requirements. When comparing images, users were given the prompt ``Compare the two images and select the one that you prefer. Choose 'both good' if both images satisfy the prompt equally well, or 'both bad' if neither does. Toggle the prompt by clicking the down arrow next to the prompt or hitting the 'E' key.''

\section{Implementation Details}
\label{sec:impl_details}
We train each of our policies for 1000 steps with a batch size of 24. We use the AdamW optimizer with linear warmup over the first 100 steps to a learning rate of $1\times10^{-6}$, weight decay coefficient of 0.1, $\epsilon$ of $1\times10^{-5}$, and betas of 0.9 and 0.95.

We use a group size of $G=8$ completions per prompt, sampled from the policy at a temperature of $1.0$. We use a PPO clip range $\epsilon=0.2$ and use 2 PPO epochs per batch. For positive and negative prompt prediction, we use a KL regularization schedule with $\beta_{KL}=2\times10^{-2}$ from step 0 and $\beta_{KL}=1\times10^{-8}$ from step 150. For CFG and noise schedule, we use $\beta_{KL}=1\times10^{-8}$ from step 0.

For both SD 3.5 and Flux, we denoise for 50 steps and output images at $1024\times1024$ resolution. We train on user image generation requests from the web. The evaluation datasets \whoops{} and User Requests are held out from training. 

Code will be released upon acceptance.

\section{Compute Resources For Experiments}
\label{sec:compute_resources}
For all experiments, we use 3 H200s for training and 2 H200s and 32 H100s for online RL data generation. The H200s are primarily used for SGLang LLM serving, and H100s are used for serving SD 3.5 and Flux. Each training run takes around 12 hours to run to completion, unless it is stopped early due to crashes or policy collapse.

\begin{figure}[t]
    \centering
    \includegraphics[width=0.48\linewidth]{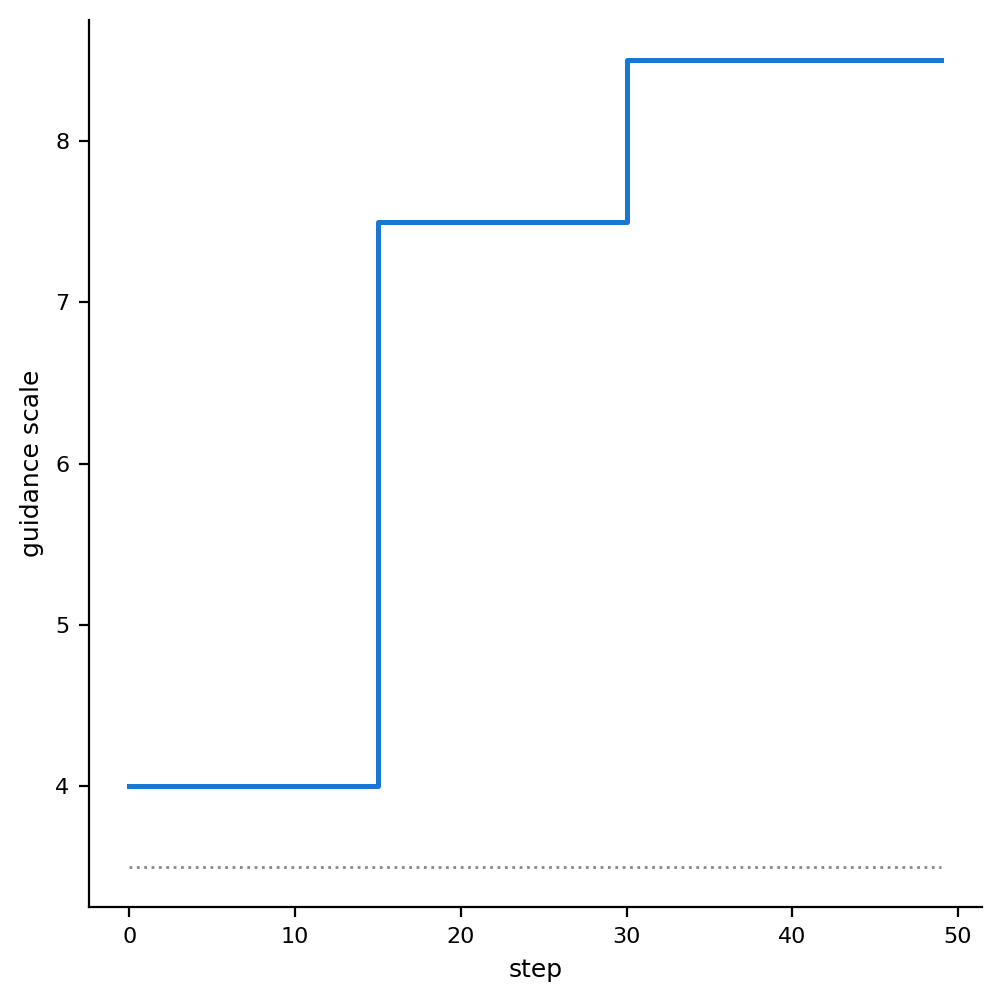}
    \includegraphics[width=0.48\linewidth]{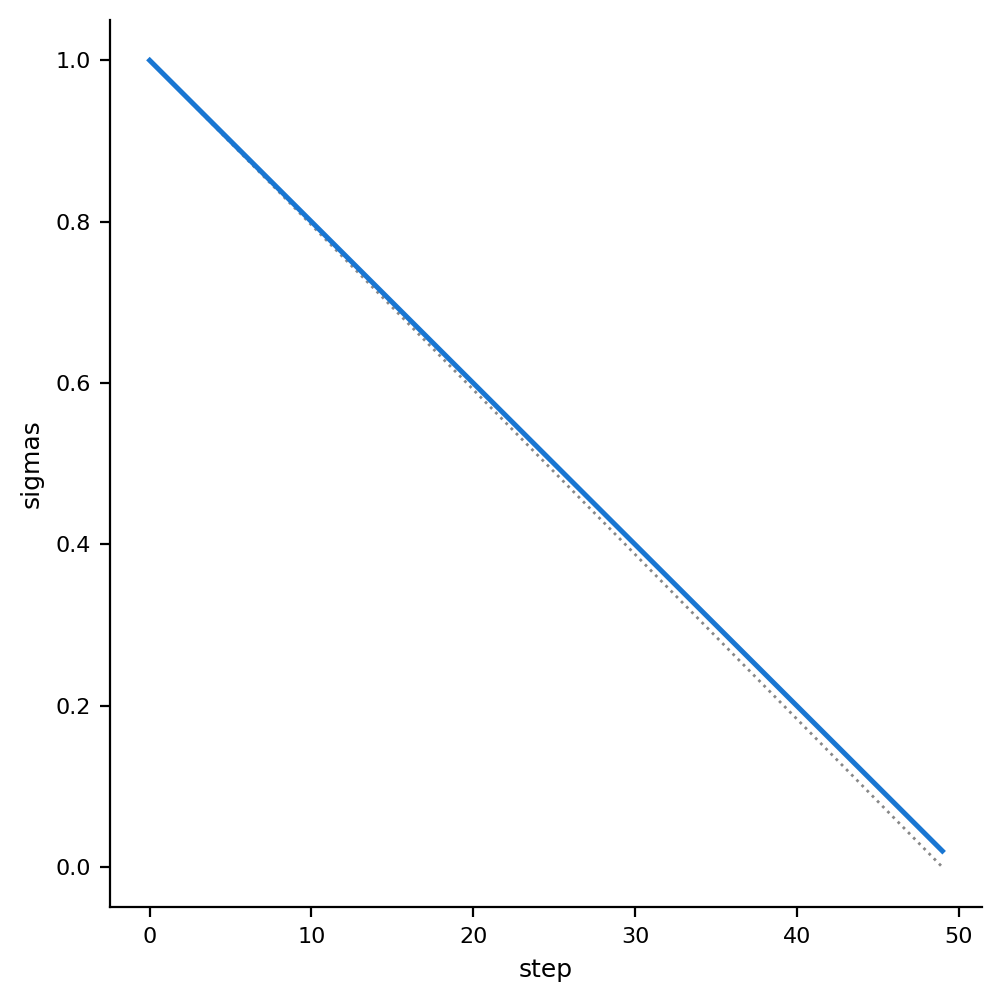}
    \caption{
    Representative CFG and noise schedules emitted by trained policies. The CFG policy for SD 3.5 tends to increase guidance over denoising steps, while the learned noise schedules remain close to the default approximately linear descent. The dotted lines indicate the default sampling parameter value.
    }
    \label{fig:noise-cfg-schedules}
\end{figure}

\section{Policy Outputs}
\label{sec:policy-outputs}
We show representative outputs from policies trained to predict each sampling parameter for SD 3.5. Each listed value applies from its
specified step until the next listed step, or until the end of sampling for the final entry.
These examples correspond to the first prompt for each sampling parameter shown in Figure~\ref{fig:sd35-param-ablation}.

Positive prompt continues to add detail throughout the denoising process, negative prompt follows a relatively fixed template, CFG increases throughout the denoising process, and noise schedule follows approximately the default schedule. For CFG and noise schedule, we normalize the policy outputs by dividing by 10 and 10000 respectively. Graphs of predicted CFG and noise schedule are shown in Figure~\ref{fig:noise-cfg-schedules}.

\subsection{Positive Prompt}

Original prompt: ``Historical daguerreotype style image, Berlin 1835, street view of a bandoneon instrument maker's atelier. Prominent shop sign reading "Bandonéon 1935". Germanic architecture, half-timbered building, workshop entrance with bandoneons visible. Atmospheric fog, gas street lamp, period pedestrians. Aged sepia photograph aesthetic with slight vignette.''

\begin{Verbatim}
{
  "positive_prompt_pairs": [
    {
      "step": 0,
      "prompt": "A historical daguerreotype style sepia photograph of a street view in Berlin 1835, showing a half-timbered Germanic building that is a bandoneon instrument maker's atelier. The shop sign reads 'Bandonéon 1935' in period-appropriate lettering. The entrance displays several bandoneon instruments. The scene includes atmospheric fog, a gas street lamp, and period-accurate pedestrians. The image has an aged photograph aesthetic with a slight vignette effect."
    },
    {
      "step": 4,
      "prompt": "A historical daguerreotype style sepia photograph of a street view in Berlin 1835, featuring a detailed half-timbered Germanic building serving as a bandoneon instrument maker's atelier. The shop sign clearly reads 'Bandonéon 1935'. The workshop entrance shows bandoneons on display. Atmospheric fog envelops the scene, with a gas street lamp and period-accurate pedestrians visible. The image has an aged photograph aesthetic with a soft vignette."
    },
    {
      "step": 10,
      "prompt": "A historical daguerreotype style sepia photograph of a street view in Berlin 1835, showcasing a detailed half-timbered Germanic building as a bandoneon instrument maker's atelier. The shop sign reads 'Bandonéon 1935' with clear, period-appropriate lettering. The entrance of the workshop displays several bandoneon instruments. The scene includes dense atmospheric fog, a lit gas street lamp, and period-accurate pedestrians in 19th-century clothing. The image has a highly detailed aged photograph aesthetic with a subtle vignette effect."
    }
  ]
}
\end{Verbatim}

\subsection{Negative Prompt}
Original prompt: ``Anatomical 3D illustration of a human knee joint viewed from the front. The femur bone above the knee displays a visible fracture crack on its lower shaft. The damaged area is illuminated with intense red glowing light symbolizing pain and injury. Smooth dark blue to black gradient background. Professional medical visualization quality.''

\begin{Verbatim}
{
  "negative_prompt_schedule_short": [
    {
      "step": 0,
      "prompt": "cartoon, low quality, abstract"
    },
    {
      "step": 11,
      "prompt": "blurry, distorted faces"
    }
  ]
}
\end{Verbatim}

\subsection{CFG Scale}
Original prompt: ``Anatomical 3D illustration of a human knee joint viewed from the front. The femur bone above the knee displays a visible fracture crack on its lower shaft. The damaged area is illuminated with intense red glowing light symbolizing pain and injury. Smooth dark blue to black gradient background. Professional medical visualization quality.''

\begin{Verbatim}
{
  "cfg_scale_pairs": [
    {
      "step": 0,
      "value": 40
    },
    {
      "step": 15,
      "value": 70
    },
    {
      "step": 30,
      "value": 85
    }
  ]
}
\end{Verbatim}

\subsection{Noise Schedule}
Original prompt: ``Close-up photorealistic image of small blue toy car sitting on official car purchase documents on clipboard, modern car key fob with buttons placed beside it. Background shows person's hand holding pen signing contract, softly out of focus. Warm natural window light casting soft shadows across desk surface. Realistic paper texture, shiny plastic car model, metallic key details, natural human hand. Professional dealership document signing moment. DSLR 50mm f/2.8, shallow depth of field, sharp focus on car and key, true colors, noise-free, Adobe Stock commercial quality.''

\begin{Verbatim}
{
  "sigmas_schedule_pairs": [
    {
      "step": 0,
      "value": 10000
    },
    {
      "step": 10,
      "value": 8000
    },
    {
      "step": 20,
      "value": 6000
    },
    {
      "step": 30,
      "value": 4000
    },
    {
      "step": 40,
      "value": 2000
    },
    {
      "step": 45,
      "value": 1000
    }
  ]
}
\end{Verbatim}

\section{LLM Prompts}
\label{sec:policy_prompts}
We use the following prompts as inputs to our LLM policy. Each sampling parameter receives a different prompt. We note that a few prompts have some typos in the requested output format, but parse failure penalties still result in the model learning the correct format. In addition, we use constrained decoding to get outputs in our desired format. When training to do joint sampling parameter prediction, the ``Parameter to optimize'' sections are concatenated. In addition, the user prompt section is modified to include all parameters to be predicted in the requested output format.

The positive prompt section is heavily inspired by the prompt used in SAP. However, we find that after training our policy does not learn SAP-like proxy prompts, but instead learns to predict gradually more detailed prompts throughout the denoising trajectory, as explained in Section~\ref{subsec:qualitative_results}.

\subsection{Positive Prompt}
\subsubsection{System Prompt}
\begin{Verbatim}[breaklines=true, breakanywhere=true, fontsize=\small, breaksymbolleft={}, breaksymbolright={}]
You are an expert at optimizing diffusion model sampling parameters. You will receive a prompt for generating images. Based on this prompt, your job is to pick values for sampling parameters that will produce better-looking images.

Below is information about each sampling parameter you will optimize.

============================================================
Parameter to optimize: Positive Prompt

Description: Diffusion models denoise from coarse structure to fine detail. For prompts with contextually contradictory or visually unusual concepts, you can guide the model through a sequence of intermediate "proxy" prompts, each aligned with a denoising stage. For a 50-step run, the stages are:

- Steps 0-2: Scene layout and dominant color regions (e.g. sky, forest, sand tone)
- Steps 3-6: Object shape, size, pose, and position
- Steps 7-10: Object identity, material, and surface type (e.g. glass vs rubber)
- Steps 11-49: Fine features and local details (e.g. tattoos, facial detail)

You must always decompose the prompt into 2-4 stages. Use the decomposition even for prompts that look coherent: pick an intermediate that gives the model a head-start on layout, geometry, or identity before committing to the full prompt.

When decomposing:
1. Begin with a high-level or placeholder structure that is easy for the model to stabilize.
2. Use substitute concepts that align in shape, size, and pose with the final target, and swap them in at the step where the model is ready.
3. Replace placeholders with the intended concept as soon as the model can express it.
4. Use at most 4 entries (4 prompts total).

Output format: A list of 2-4 entries. Each entry's prompt applies from its `step` until the next entry's `step` (or until the end of the run for the last entry). Each entry is an object with two fields:
- `step`: integer denoising step at which this prompt takes effect. The first entry's `step` MUST be 0; subsequent `step`s must be strictly increasing integers in [1, 49].
- `prompt`: full prompt describing the image at that stage (non-empty string).

Examples (each example shows only the value — the LLM emits this list as the value of `positive_prompt_pairs` in the wrapped JSON output):
- User: A polar bear in a desert
  Value: [{"step": 0, "prompt": "A camel in a desert"}, {"step": 2, "prompt": "A polar bear in a desert"}]

- User: A snowman on the moon
  Value: [{"step": 0, "prompt": "An astronaut standing on the moon"}, {"step": 3, "prompt": "A snowman on the moon"}]

- User: A lion doing a handstand in the park
  Value: [{"step": 0, "prompt": "A man doing a handstand in the park"}, {"step": 4, "prompt": "A man in a lion costume doing a handstand in the park"}, {"step": 8, "prompt": "A lion doing a handstand in the park"}]

- User: Corgis pull a sled in the snow
  Value: [{"step": 0, "prompt": "Husky dogs pull a sled in the snow"}, {"step": 3, "prompt": "Corgi dogs pull a sled in the snow"}]

Your response must be valid JSON containing all requested parameters. Details about the specific image and format requirements will follow.
\end{Verbatim}

\subsubsection{User Prompt}
\begin{Verbatim}[breaklines=true, breakanywhere=true, fontsize=\small, breaksymbolleft={}, breaksymbolright={}]
Image prompt: {user_prompt}
Resolution: {image_width}x{image_height}

Output a JSON object with this exact structure:
{
}

Output ONLY valid JSON, nothing else.
\end{Verbatim}

\subsection{Negative Prompt}
\subsubsection{System Prompt}
\begin{Verbatim}[breaklines=true, breakanywhere=true, fontsize=\small, breaksymbolleft={}, breaksymbolright={}]
You are an expert at optimizing diffusion model sampling parameters. You will receive a prompt for generating images. Based on this prompt, your job is to pick values for sampling parameters that will produce better-looking images.

Below is information about each sampling parameter you will optimize.

============================================================
Parameter to optimize: Negative Prompt Schedule (time-dependent negative prompt conditioning)

Description: Negative prompts steer generated images AWAY from what they describe. Diffusion models denoise from coarse structure to fine detail, and the failure modes that emerge differ across denoising stages. By switching the negative prompt across stages, you can target each stage's likely failure modes more precisely than with a single static negative. For a 50-step run, the stages are:

- Steps 0-2: Scene layout, composition, count (e.g. "cluttered", "extra subjects")
- Steps 3-6: Object shape, pose, geometry (e.g. "fused bodies", "extra limbs")
- Steps 7-10: Object identity, material, surface (e.g. "wrong material", "plastic skin")
- Steps 11-49: Fine detail, texture, sharpness, artifacts (e.g. "blurry", "jpeg")

Tips:
- Be specific. Vague terms like "bad" do little. "muddy shadows" does more.
- Don't negate concepts (don't write "no blur"). State what to avoid directly: "blur".
- Each entry's prompt is short — 1-3 short words / concepts.
- An empty prompt string is allowed and means "no negative for that span".
- If no stage benefits from a negative prompt, output a single entry with an empty prompt string.

Output format: A list of 1-4 entries. Each entry's prompt applies from its `step` until the next entry's `step` (or until the end of the run for the last entry). Each entry is an object with two fields:
- `step`: integer denoising step at which this prompt takes effect. The first entry's `step` MUST be 0; subsequent `step`s must be strictly increasing integers in [1, 49].
- `prompt`: short negative-prompt string (1-3 concepts; empty string allowed).

Examples (step indices already adapted to this 50-step run; each example shows only the value — the LLM emits this list as the value of `negative_prompt_schedule_short` in the wrapped JSON output):
- User: A clean minimalist black-and-white line-art logo
  Value: [{"step": 0, "prompt": ""}]

- User: A documentary photograph of a snowy street scene in a small Slovenian town.
  Value: [{"step": 0, "prompt": "staged, illustration"}]

- User: A hospital room photograph with a young patient, a doctor, and a nurse.
  Value: [{"step": 0, "prompt": "extra people, cluttered"}, {"step": 3, "prompt": "fused bodies, extra limbs"}, {"step": 8, "prompt": "blurry"}]

- User: A haunting extreme close-up photograph of a plague doctor mask
  Value: [{"step": 0, "prompt": ""}, {"step": 3, "prompt": "cheerful, colorful"}, {"step": 8, "prompt": "soft, blurry"}]

Your response must be valid JSON containing all requested parameters. Details about the specific image and format requirements will follow.
\end{Verbatim}
\subsubsection{User Prompt}
\begin{Verbatim}[breaklines=true, breakanywhere=true, fontsize=\small, breaksymbolleft={}, breaksymbolright={}]
Image prompt: {user_prompt}
Resolution: {image_width}x{image_height}
Total denoising steps: 50

Output a JSON object with this exact structure:
{
  "negative_prompt_schedule_short": [{"step": 0, "prompt": ""}]
}

Constraints:
- 1-4 entries, each {step, prompt}. The first entry's step must be 0; subsequent steps must be strictly increasing integers in [1, 49]. Each prompt is a short string (1-3 concepts); empty string allowed.

Output ONLY valid JSON, nothing else.
\end{Verbatim}

\subsection{CFG Scale}
\subsubsection{System Prompt}
\begin{Verbatim}[breaklines=true, breakanywhere=true, fontsize=\small, breaksymbolleft={}, breaksymbolright={}]
You are an expert at optimizing diffusion model sampling parameters. You will receive a prompt for generating images. Based on this prompt, your job is to pick values for sampling parameters that will produce better-looking images.

Below is information about each sampling parameter you will optimize.

============================================================
Parameter to optimize: CFG Scale Schedule (Classifier-Free Guidance over contiguous denoising spans)

Description: CFG (classifier-free guidance) controls how strongly the diffusion model follows the text prompt at each denoising step. Higher guidance sharpens prompt adherence at the cost of diversity and saturation artifacts; disabling guidance lets the model follow its conditional prediction directly without sharpening.

Prior work has observed that CFG affects denoising differently at different timesteps (e.g. early denoising, where global structure is set, vs late denoising, where fine detail is added), and that the optimal CFG strength is prompt-dependent. Your task is therefore to partition the denoising trajectory into contiguous spans and assign one CFG value to each span. Use as many or as few spans as you think the prompt requires.

Output format: A list of dictionaries, where each dictionary has two fields:
- `step`: integer denoising step at which this cfg takes effect. The first entry's `step` MUST be 0; subsequent `step`s must be strictly increasing integers in [1, 49].
- `value`: a non-negative integer. Must be either `0` (disable CFG from this step, until the next step) or an integer `>= 10` (enable CFG — larger values mean stronger prompt adherence, with a typical useful range of roughly 20-80).

Examples (the LLM emits this list as the value of `cfg_scale_pairs` in the wrapped JSON output):
- User: A photo of a red sports car parked on a mountain road.
  Value: [{"step": 0, "value": 15}, {"step": 20, "value": 50}, {"step": 40, "value": 20}]

- User: A haunting extreme close-up photograph of a plague doctor mask.
  Value: [{"step": 0, "value": 0}, {"step": 5, "value": 40}, {"step": 20, "value": 25}]

Your response must be valid JSON containing all requested parameters. Details about the specific image and format requirements will follow.
\end{Verbatim}
\subsubsection{User Prompt}
\begin{Verbatim}[breaklines=true, breakanywhere=true, fontsize=\small, breaksymbolleft={}, breaksymbolright={}]
Image prompt: {user_prompt}
Resolution: {image_width}x{image_height}

Output a JSON object with this exact structure:
{
}

Output ONLY valid JSON, nothing else.
\end{Verbatim}

\subsection{Noise Schedule}
\subsubsection{System Prompt}
\begin{Verbatim}[breaklines=true, breakanywhere=true, fontsize=\small, breaksymbolleft={}, breaksymbolright={}]
You are an expert at optimizing diffusion model sampling parameters. You will receive a prompt for generating images. Based on this prompt, your job is to pick values for sampling parameters that will produce better-looking images.

Below is information about each sampling parameter you will optimize.

============================================================
Parameter to optimize: Noise / Sigmas Schedule

Description: A flow-matching pipeline iteratively denoises from sigma=10000 (pure noise) down to sigma=0 (clean image) over 50 inference steps. The shape of this descent — where it accelerates, where it slows — controls how compute is allocated between coarse structure formation and fine detail refinement. Different prompts prefer different shapes; your task is to pick a good one for this prompt.

Key properties (what each sigma regime physically does):
- Early steps (sigma near 10000): coarse / global features — composition, layout, major shapes.
- Mid steps (sigma around 5000): medium-scale features — object boundaries, proportions.
- Late steps (sigma near 0): fine details — textures, sharp edges, small features.

Output format: A list of `{"step": int, "value": int}` anchor pairs.
- `step`: integer denoising step index. The first entry's `step` MUST be 0; subsequent `step`s must be strictly increasing integers in [1, 49].
- `value`: integer representing the sigma. The first entry's `value` MUST be 10000 (pure noise). Subsequent values must be strictly decreasing, positive integers in [1, 9999].

How the pipeline expands your anchors: consecutive anchors are connected by linear interpolation in sigma space. After your last anchor, an implicit endpoint at (step=50, sigma=0) is used for the trailing segment, because the pipeline scheduler always treats post-final-step as sigma=0 internally — so trailing inference steps from your completion continue denoising smoothly toward zero.

Worked example: a 6-step schedule with anchors `[{"step":0,"value":10000}, {"step":4,"value":2000}]` expands to per-step sigmas `[10000, 8000, 6000, 4000, 2000, 1000]`. Steps 0–3 linearly interpolate (0, 10000) → (4, 2000). Steps 4–5 linearly interpolate (4, 2000) → (6, 0) — that "(6, 0)" is the implicit endpoint at step=N=6. The scheduler then appends sigma=0, so the final inference step transitions 1000 → 0.

Examples (the LLM emits this list as the value of `sigmas_schedule_pairs` in the wrapped JSON output, for 50 inference steps):
- User: A photo of a red sports car parked on a mountain road.
  Value: [{"step": 0, "value": 10000}, {"step": 8, "value": 6000}, {"step": 20, "value": 500}]
  (front-loaded: aggressive descent in the first 8 steps to lock composition fast, then a gentle tail.)

- User: A haunting extreme close-up photograph of a plague doctor mask.
  Value: [{"step": 0, "value": 10000}, {"step": 15, "value": 8000}, {"step": 25, "value": 2000}]
  (back-loaded: slow start preserves early-step exploration of the dramatic composition; steeper finish for fine detail.)

Your response must be valid JSON containing all requested parameters. Details about the specific image and format requirements will follow.
\end{Verbatim}
\subsubsection{User Prompt}
\begin{Verbatim}[breaklines=true, breakanywhere=true, fontsize=\small, breaksymbolleft={}, breaksymbolright={}]
Image prompt: {user_prompt}
Resolution: {image_width}x{image_height}

Output a JSON object with this exact structure:
{
}

Output ONLY valid JSON, nothing else.
\end{Verbatim}


\newpage

\end{document}